%% file: paper.tex
\PassOptionsToPackage{table}{xcolor}
\documentclass[]{fairmeta}
\usepackage[numbers]{natbib}
\usepackage{amsthm,amsmath,amssymb}
\usepackage{graphicx}
\usepackage{tikz}
\usetikzlibrary{arrows.meta,positioning}
\usepackage{booktabs}
\usepackage[table]{xcolor}
\usepackage{bm}
\usepackage{xspace}
\usepackage{tabularx}
\usepackage{geometry}
\usepackage{multirow}
\usepackage{enumitem}
\usepackage{placeins}
\usepackage{iftex}
\ifPDFTeX
\else
  \usepackage{fontspec}
\fi
\usepackage{titlesec}
\usepackage{subcaption}
\usepackage[hidelinks]{hyperref}
\newif\ifshowthreepoint
\showthreepointfalse

\newcolumntype{L}{>{\raggedright\arraybackslash}X}
\newcolumntype{C}{>{\centering\arraybackslash}X}
\newcolumntype{R}{>{\raggedleft\arraybackslash}X}

\ifPDFTeX
  \newcommand{\calmfont}[1]{{\mdseries #1}}
  \titleformat*{\section}{\Large\mdseries}
  \titleformat*{\subsection}{\large\mdseries}
  \titleformat*{\subsubsection}{\normalsize\mdseries}
\else
  \newfontfamily\calmregular[FakeBold=2]{Outfit-Regular.ttf}
  \newcommand{\calmfont}[1]{{{\calmregular\mdseries #1}}}
  \titleformat*{\section}{\Large\mdseries\calmregular}
  \titleformat*{\subsection}{\large\mdseries\calmregular}
  \titleformat*{\subsubsection}{\normalsize\mdseries\calmregular}
\fi

\newtheorem{theorem}{Theorem}[]

\newtheorem{remark1}[theorem]{Remark}

\newcommand{\modelname}{HoloMotion-2\xspace}
\definecolor{reportblue}{HTML}{234D78}
\definecolor{reportlight}{HTML}{EDF4FA}
\definecolor{reportgreen}{HTML}{245F55}
\definecolor{reportmint}{HTML}{EDF6F2}

\title{\calmfont{HoloMotion-2 Technical Report}}
\author{\calmfont{Liu Liu\textsuperscript{\ensuremath{*}}{\normalfont\rmfamily,} Xihan Ma{\normalfont\rmfamily,} Lulu Cao{\normalfont\rmfamily,} Xuyuan Miao{\normalfont\rmfamily,} Bo Zhang{\normalfont\rmfamily,} Zhengchen Wu{\normalfont\rmfamily,} Min Du{\normalfont\rmfamily,} Kaihui Wang{\normalfont\rmfamily,} Yucheng Wang\textsuperscript{\ensuremath{*}}{\normalfont\rmfamily,} Zhizhong Su\textsuperscript{\ensuremath{\dagger}}}}
\affiliation{\calmfont{\mdseries Horizon Robotics HoloMotion Team}}
\contribution[\dagger]{Corresponding author.\quad\textsuperscript{\ensuremath{*}}Project lead.}

\metadata[Code]{\href{https://github.com/HorizonRobotics/HoloMotion}{HoloMotion}}
\metadata[Model]{\href{https://huggingface.co/HorizonRobotics}{Model weights on Hugging Face}}
\webpage{\href{https://horizonrobotics.github.io/robot_lab/holomotion_2}{horizonrobotics.github.io/robot\_lab/holomotion\_2}}
\correspondence{Zhizhong Su at \email{zhizhong.su@horizon.auto}}

\hypersetup{pdftitle={HoloMotion-2 Technical Report}}
\begin{document}

\abstract{\input{sections/00_abstract}
}
\maketitle

\input{sections/01_introduction}
\input{sections/02_background}
\input{sections/03_method}
\input{sections/05_experiments}
\input{sections/05_application}
\input{sections/06_conclusion}
\input{sections/07_future_work}

{
\linespread{0.97}\selectfont
\setlength{\bibsep}{2pt plus 0.3pt minus 0.3pt}
\bibliographystyle{unsrtnat}
\bibliography{paper}
}

\input{sections/08_appendix}

\input{sections/09_policy_interface}
\input{sections/10_reward_terms}
\input{sections/11_training_schedule}

\end{document}

%% file: sections/00_abstract.tex

Achieving general-purpose humanoid control requires moving beyond rigid, single-format tracking interfaces to serve diverse upstream systems, from teleoperators and motion planners to high-level embodied agents. This calls for a universal whole-body controller capable of executing \emph{any command}---spanning spatially and temporally sparse pose targets, task-level goals, and ultimately multimodal instructions such as natural language. However, mapping such under-specified commands to physically viable whole-body motion is severely under-constrained and multi-modal. Toward this vision, we present \textbf{\modelname}, a generative behavior foundation model trained in two stages. First, we train a sparse prior and latent-routed Mixture-of-Experts (MoE) state--action flow policy via behavior cloning and DAgger, distilling dense demonstrations to handle spatio-temporally sparse commands. Second, we perform physical post-training via closed-loop RL with a differentiable world model---eliminating sample-likelihood estimation by backpropagating multi-step reward gradients directly through the learned dynamics and the flow solver. As a result, a single controller seamlessly executes both dense motion sequences and sparse task commands. Across major general motion tracking benchmarks---including HumanTracker and MotionChallenge---world-model optimization improves tracking success by over 50\% over supervised imitation and achieves state-of-the-art performance over published baselines. Real-world deployments on Unitree G1 across diverse downstream tasks further demonstrate its physical robustness and generalization. \mbox{Our code and model will be released soon.}

%% file: sections/01_introduction.tex
\section{Introduction}
\label{sec:introduction}

Over the past year, advances in reinforcement learning (RL), human motion imitation, and large-scale physics simulation have propelled rapid progress in general motion tracking (GMT), enabling humanoid controllers to execute increasingly diverse motor skills \citep{beyondmimic,sonic,holomotion1,scalebfm}. Alongside algorithmic advances, the rapid expansion of human and embodied motion datasets has further broadened the movement repertoire accessible to modern trackers \citep{amass,motionmillion,motiondecode,bonesseed,snapmogen,humanoideveryday}. Yet existing controllers remain largely fragmented around specialist trackers or dense full-body playback: a tracker tuned for a narrow dynamic repertoire cannot reliably cover everyday physical tasks such as bending down to pick up trash or opening a drawer. Moving humanoid control from controlled demonstrations toward practical real-world deployment requires replacing rigid, single-format tracking interfaces with a universal whole-body behavior foundation model (BFM). To this end, \modelname aims to build a unified controller capable of executing \emph{any command}---spanning complete motion references, spatially and temporally sparse pose targets, task-level goals, and ultimately high-level multimodal instructions such as natural language.


Supporting such flexible interfaces enables a clean division of labor between a high-level ``brain'' and a low-level ``cerebellum'' in embodied AI. In current humanoid loco-manipulation pipelines, the brain is severely overloaded: while already handling long-horizon task decomposition, reasoning, and 3D spatial understanding, it is also forced to generate \emph{high-frequency, dense, physically accurate} whole-body trajectories, creating a fragile bottleneck. Adopting sparse commands relieves this burden with three key advantages. First, physical tasks are inherently sparse: they specify only key end-effector targets at critical moments, leaving intermediate postures and footwork unconstrained \citep{metatron}. Second, delegating balance and stepping to the cerebellum improves robustness in dynamic or cluttered environments, enabling autonomous stance adjustment, perturbation recovery, and agile interactions such as boxing. Third, sparse targets unify high-level action spaces across embodiments, bridging high-DoF humanoids with lower-dimensional platforms such as wheeled or tabletop manipulators.

Moving from dense tracking to sparse commands fundamentally changes the learning problem. Since sparse or abstract commands are under-constrained, a single target admits multiple physically valid solutions. Deterministic regression averages across these incompatible modes, often erasing the balance-critical weight shifts needed to remain stable. Conditional flow matching resolves this ambiguity by learning a continuous vector field that maps noise samples to coherent, multi-mode trajectory distributions \citep{flowmatching}. While flow-based policies have proven effective in manipulation and autonomous driving \citep{pi0,goalflow}, humanoid whole-body control introduces a distinct challenge: the controller must jointly synthesize both the uncommanded body motion and the low-level motor actions required to realize it physically.

Even with an expressive generative architecture, pure imitation learning and open-loop flow distillation remain insufficient for reliable whole-body control. Under sparse conditioning, small per-step discrepancies compound into out-of-distribution states during closed-loop execution, causing high joint jerk, tracking drift, and physical instability. Refining iterative flow policies with closed-loop RL is itself challenging because multi-step sampling lacks a tractable action likelihood. Recent approaches circumvent this in visual generation by scoring sampled outputs with group-relative rewards---such as Flow-GRPO \citep{flowgrpo} and DanceGRPO \citep{dancegrpo}---and in robot control by injecting learned sampling noise (ReinFlow \citep{reinflow}) or optimizing surrogate flow-matching objectives (FPO \citep{lyufpo}). However, these model-free methods remain constrained by high sample complexity when scaling across diverse, contact-rich humanoid behaviors. Differentiable world models learned from physics simulation, as demonstrated by SuperMimic \citep{supermimic} for humanoid control, offer a principled alternative: by providing a differentiable physics surrogate, they bypass explicit sample-likelihood estimation altogether and backpropagate analytical multi-step reward gradients directly into the policy. Extending this physical grounding to a generative BFM, however, requires propagating gradients seamlessly through both the learned dynamics and an iterative flow solver.

To address this challenge, we present \textbf{\modelname}, a generative behavior foundation model built through a two-stage pipeline that couples scalable imitation with closed-loop physical refinement. First, supervised dense-to-sparse flow distillation via behavior cloning and iterative dataset aggregation (DAgger) \citep{dagger,maskedmimic,flowmimic} trains a transformer sparse prior paired with a latent-routed Mixture-of-Experts (MoE) state--action flow decoder, converting dense tracking expertise into a unified sparse command interface. Second, building on differentiable world-model policy optimization \citep{supermimic}, we refine the policy end-to-end against imagined closed-loop rollouts, backpropagating analytical multi-step tracking and smoothness gradients through both the learned dynamics and the iterative flow solver.

Our main contributions are threefold:
\begin{enumerate}[leftmargin=*,nosep]
  \item \textbf{Unified Generative Framework for Dense-to-Sparse Control:} We establish a generative BFM---pairing a transformer sparse prior with a latent-routed MoE joint state--action flow decoder---that unifies spatio-temporal sparse commands and resolves multi-mode ambiguity as a versatile System~0 controller for diverse upstream interfaces.
  \item \textbf{Closed-Loop World-Model RL for Generative Policies:} We further substantially improve tracking success  and physical stability over 50\% through closed-loop world-model RL---propagating multi-step gradients directly through learned dynamics and the flow solver---establishing a general paradigm for physical post-training of generative policies.
  \item \textbf{SOTA Benchmarks and Cross-Scenario Real-World Validation:} \modelname sets state-of-the-art results on large-scale GMT benchmarks (HumanTracker and MotionChallenge) and sparse-command evaluations, supported by systematic ablations, and transfers reliably across diverse real-world tasks on a Unitree G1.
\end{enumerate}

%% file: sections/02_background.tex
\section{Related Work}
\label{sec:background}

\paragraph{Whole-body tracking and humanoid controllers.}
Recent humanoid controllers have rapidly expanded both motion diversity and real-world transferability. BeyondMimic demonstrates hardware transfer of dynamically rich reference motions \citep{beyondmimic}, SONIC scales data, compute, and model capacity for teleoperation and loco-manipulation \citep{sonic}, and ScaleBFM systematically investigates world-frame tracking, rollout scale, and policy capacity \citep{scalebfm}. Extending tracking to specialized settings, HEFT incorporates privileged reference guidance and a payload curriculum for heavy-payload teleoperation \citep{heft}, MimicAll generalizes across distinct humanoid embodiments, and MimicLite and GRIT provide open implementations for tracking and teleoperation \citep{mimicall,mimiclite,grit}. Supporting this rapid progress in general motion tracking (GMT), the growing scale of human and embodied motion datasets---spanning foundational motion-capture archives \citep{amass,lafan1,omomo} alongside recent million-scale, expressive, and robot-ready corpora \citep{motionmillion,motiondecode,bonesseed,snapmogen,humanoideveryday}---has substantially enriched the motion distributions available for training. Generative video also provides a source of motion data. Uni-Mo reconstructs three-dimensional quadruped trajectories from synthesized videos to train tracking policies \citep{unimo}. Despite these advances, most trackers still assume that an upstream module supplies dense, frame-by-frame full-body references---or rely on a separate external motion generator---leaving the high-level planner burdened with detailed trajectory synthesis. 

\paragraph{Sparse-command control and generative robotics policies.}
Flow matching provides a way to generate continuous actions and trajectories conditioned on a task. For robotic manipulation, $\pi_0$ and LingBot-VLA use it to generate action sequences from images and language instructions \citep{pi0,lingbotvla}. For autonomous driving, GoalFlow uses it to generate vehicle trajectories conditioned on candidate goals \citep{goalflow}. Humanoid whole-body control introduces an additional requirement: the generated actions must maintain balance as contacts change. Within humanoid robotics, several efforts formulate control as conditional generation from partial specifications: MaskedMimic performs physics-based motion completion from masked goals via a latent prior distilled from a privileged encoder \citep{maskedmimic}, METATRON feeds timed body-part targets to a frozen generative base policy for programmable skills \citep{metatron}, FlowMimic applies action-space flow distillation over masked targets \citep{flowmimic}, PredActor jointly predicts future states and actions from proprioception for trajectory-level guidance \citep{predactor}, and MotionBricks composes reusable motion primitives \citep{motionbricks}. Yet relying solely on open-loop supervised distillation, frozen priors, or kinematic primitive composition struggles to ensure closed-loop physical stability and smoothness under sparse conditioning. \modelname addresses this gap by pairing a timed sparse prior with a joint state--action MoE flow decoder and optimizing the full decoding path end-to-end through closed-loop differentiable rollouts.


\paragraph{Closed-loop RL and differentiable world-model optimization.}
Improving the efficiency of closed-loop whole-body reinforcement learning is critical for scaling humanoid controllers. SuperMimic pioneers replacing non-differentiable physics simulators with a learned differentiable world model that amortizes physical dynamics across diverse reference motions, enabling policy optimization via backpropagation through short imagined rollouts \citep{supermimic}. Building on this paradigm, subsequent works further optimize training throughput and stability: InstantMimic and InstantTrack reduce GPU overhead through execution and optimizer refinements \citep{instantmimic,instanttrack}, BlinkMimic and BlinkMimic~2 leverage failure-directed resets and long-horizon rollouts to lower iteration costs \citep{blinkmimic,blinkmimic2}, and XMimic co-compiles simulation and learning across diverse locomotion and tracking regimes \citep{xmimic}. Beyond accelerating deterministic dense trackers, we identify a broader role for world-model differentiability: propagating analytical physical gradients directly into expressive generative motion predictors and planners to elevate the physical executability of synthesized trajectories. Specifically, \modelname differentiates end-to-end through both learned dynamics and the iterative flow integration steps, grounding its sparse prior and joint state--action flow policy in closed-loop physics.

%% file: sections/03_method.tex
\section{Method}
\label{sec:method}


Figure~\ref{fig:pipeline} presents an overview of \modelname, a unified generative framework for dense and spatio-temporal sparse humanoid control. We begin by formalizing the control problem, then introduce our policy architecture, and finally detail our two-stage training pipeline.

\subsection{Problem Formulation}
\label{sec:problem-formulation}

Our goal is to learn a unified whole-body controller $\pi$ that executes arbitrary dense or sparse motion commands while maintaining physical balance. At control step $t$, an upstream system specifies \emph{which body parts should reach which poses, and when} via timed pose constraints:
\begin{equation}
  C_t = \bigl\{(\ell,\,\tau-t,\,\widehat T_{\ell,\tau})
  \,\big|\, \tau\in\mathcal T_t,\ \ell\in M_\tau\bigr\},
  \label{eq:sparse-command}
\end{equation}
where $\mathcal T_t$ denotes the commanded future timestamps (with remaining time-to-go $\tau - t \ge 0$), $M_\tau$ is the constrained link subset at time $\tau$, and $\widehat T_{\ell,\tau}$ is the target pose of link $\ell$ in the robot's current local frame. Standard dense tracking is recovered when $M_\tau$ spans the full body over consecutive frames, whereas subsampling $M_\tau$ or spacing out $\mathcal T_t$ yields spatially and temporally sparse commands.

Let $s_t$ denote the robot's physical state and $h_t$ a window of recent proprioceptive observations available to the controller. At each control tick, conditioned on $(h_t, C_t)$, the policy samples a short-horizon trajectory $\mathbf X_t$ of paired future whole-body states and joint actions, executing only the first action $a_t$ in a receding-horizon closed loop:
\begin{equation}
  \mathbf X_t = \bigl\{(\tilde s_{t+i}, a_{t+i})\bigr\}_{i=0}^{H-1} \sim \pi(\,\cdot \mid h_t, C_t),
  \qquad
  s_{t+1} \sim \mathcal P(\,\cdot \mid s_t, a_t),
  \label{eq:control-step}
\end{equation}
where $\tilde s_{t+i}$ and $a_{t+i}$ denote the predicted whole-body state and joint action at horizon offset $i$, and $\mathcal P$ represents the environment transition dynamics. The control objective is to satisfy every timed pose constraint in $C_t$ while preserving whole-body balance, physical feasibility, and motion smoothness across all unconstrained links and intermediate frames.

This formulation reflects two key design choices for sparse control. First, since a sparse command $C_t$ admits multiple valid stepping and balance strategies, modeling $\pi$ as a generative distribution avoids averaging across conflicting motion modes. Second, predicting future whole-body states $\tilde s_{t+i}$ alongside joint actions $a_{t+i}$ guides the controller to plan coherent full-body motion across unconstrained joints. 

\input{figure/training_pipeline}

\subsection{Sparse Prior and Latent-Routed Flow Decoder}
\label{sec:architecture}

\paragraph{Sparse prior and privileged posterior.}
To encode arbitrary command masks into a fixed 64-dimensional conditioning space, the sparse prior $p_\phi$ processes the current observation, age-embedded proprioceptive history frames, and up to five masked future goal slots via an $N$-layer transformer encoder ($N\in\{3,6\}$), outputting a diagonal Gaussian from the current-observation token. During supervised distillation, a privileged posterior $q_\psi$ additionally observes dense future targets $C_t^{\mathrm{dense}}$ and predicts a residual mean shift $\Delta\mu_\psi$; sampled latents are $\ell_2$-normalized before decoding to bound scale:
\begin{equation}
  p_\phi(z\mid h_t, C_t) = \mathcal N\!\bigl(\mu_\phi,\mathrm{diag}(\sigma_\phi^2)\bigr), \quad
  q_\psi(z\mid h_t, C_t^{\mathrm{dense}}, C_t) = \mathcal N\!\bigl(\mu_\phi + \Delta\mu_\psi,\mathrm{diag}(\sigma_\psi^2)\bigr), \quad
  \hat z = \frac{z}{\|z\|_2}.
  \label{eq:prior-posterior}
\end{equation}
At inference time, only $p_\phi$ is active, and deterministic inference feeds its $\ell_2$-normalized mean $\hat\mu_\phi = \mu_\phi / \|\mu_\phi\|_2$ directly to the decoder.

\paragraph{Latent-routed MoE flow decoder.}
Within a single control tick, the decoder predicts a velocity field $v_\theta(x^{(j)}, u_j, h_t, \hat z)$ over noisy normalized state--action sequences. To scale skill capacity without hurting control frequency, its core Mixture-of-Experts (MoE) layer pairs one always-active shared expert $F_{\mathrm{shared}}$ (for universal balance dynamics) with two experts selected from $E=1{,}024$ routed experts solely by the $\ell_2$-normalized command latent $\hat z$:
\begin{equation}
  \mathrm{MoE}(y;\hat z) = F_{\mathrm{shared}}(y) + \sum_{e\in\mathcal I(\hat z)} \alpha_e(\hat z)\,F_e(y),
  \qquad \mathcal I(\hat z) = \mathrm{Top}\text{-}2\bigl(g(\hat z)\bigr),
  \label{eq:moe}
\end{equation}
where $y$ denotes intermediate decoder features, $g(\hat z)$ produces router logits, and $\alpha_e(\hat z)$ are the corresponding routing weights. Starting from Gaussian noise $x^{(0)} = x_0 \sim \mathcal N(0,I)$, the decoder integrates the normalized trajectory over $K$ uniform Euler steps ($K=5$, $u_j = j/K$):
\begin{equation}
  x^{(j+1)} = x^{(j)} + \frac{1}{K}\,v_\theta\!\bigl(x^{(j)}, u_j, h_t, \hat z\bigr), \qquad j = 0,\dots,K-1.
  \label{eq:euler}
\end{equation}
Crucially, since $(h_t, \hat z)$ remain fixed throughout a control tick, both the conditioning embeddings and the MoE routing decisions $\bigl(\mathcal I(\hat z), \alpha(\hat z)\bigr)$ in Eq.~\eqref{eq:moe} are computed once and reused across all $K$ Euler steps. The denoised sequence $x^{(K)}$ is then denormalized to obtain $\mathbf X_t$, from which the first joint action $a_t$ is extracted for execution.

\FloatBarrier
\subsection{Dense-to-Sparse Flow Distillation}
\label{sec:distillation}

\paragraph{Behavior cloning.}
We roll out a dense motion-tracking teacher in simulation to collect normalized expert state--action sequences $x_1$. To construct sparse inputs $C_t$, we combine a ScaleBFM-style spatial link mask \citep{scalebfm} with one of three temporal profiles: \emph{dense} (20\%, five consecutive future frames), \emph{intermediate} (60\%, slots spaced 2--5 frames apart with random masking), or \emph{sparse} (20\%, five candidate slots spaced 6--10 frames apart with one active slot). All slots respect reference boundaries, holding each active target until expiration with its countdown $\tau-t$ decremented every tick.

We train the conditional flow via straight-line interpolation $x_u = (1-u)x_0 + u x_1$ with $x_0\sim\mathcal N(0,I)$ and $u\in[0,1]$ \citep{flowmatching}. Drawing $z_q\sim q_\psi$ and $\ell_2$-normalizing $\hat z_q = z_q/\|z_q\|_2$, the supervised objective is
\begin{equation}
\mathcal L_{\mathrm{sup}} =
\mathbb E\!\left[\left\|v_\theta(x_u,u,h_t,\hat z_q)-(x_1-x_0)\right\|^2_{\mathrm{valid}}\right]
+\beta\,\mathbb E\!\left[D_{\mathrm{KL}}(q_\psi\Vert p_\phi)\right]
+\eta\,\mathbb E\!\left[(\|\mu_\phi\|_2-1)^2\right],
\label{eq:flow}
\end{equation}
where the flow loss masks invalid boundary steps and averages state and action channels separately, $\beta$ aligns the deployable prior $p_\phi$ to the privileged posterior $q_\psi$, and $\eta$ prevents scale drift of $\mu_\phi$ prior to normalization.

\paragraph{Dataset aggregation.}
Since pure behavior cloning suffers from closed-loop covariate shift, we run iterative DAgger \citep{dagger}. In each round, the student executes using its own prior $p_\phi$; from student-visited states, the dense teacher takes over for short recovery segments to generate corrective labels, which are aggregated into the replay buffer to re-optimize Eq.~\eqref{eq:flow}.

\subsection{Differentiable World-Model Policy Optimization}
\label{sec:worldmodel}

Open-loop flow distillation alone cannot eliminate compounding closed-loop errors or guarantee that generated motions remain physically executable under contact-rich dynamics. To ground the generative policy in closed-loop physics, we adopt differentiable world-model policy optimization \citep{supermimic}, alternating between simulator rollouts, dynamics model fitting, and end-to-end actor--critic updates. Given state $s_t$, action $a_t$, and physical context $c_t$ (encoding randomized flat-ground parameters), a learned generalized-coordinate transition model $f_\xi$ predicts body-frame velocity increments integrated into the next root and joint state. We train $f_\xi$ on uninterrupted simulator trajectories via multi-step losses over root pose, joint positions, and joint velocities.

For each actor update, we seed a horizon-$L$ imagined rollout from a replay state ($\hat s_t = s_t$) with the world model $f_\xi$ frozen. At each imagined step $k\in\{0,\dots,L-1\}$, we update the proprioceptive history $\hat h_{t+k}$, transform the upcoming sparse targets into the predicted local frame to form $\hat C_{t+k}$, and condition the flow policy on the $\ell_2$-normalized mean $\hat\mu_{\phi,t+k}$ of the sparse prior without latent sampling (matching deterministic inference). Starting from freshly sampled flow noise $x_k^{(0)}\sim\mathcal N(0,I)$, the $K$-step Euler flow policy $\pi_\theta$ integrates Eq.~\eqref{eq:euler}---followed by denormalization and first-action extraction---to produce the executed action $a_{t+k}$, and $f_\xi$ advances the imagined state:
\begin{equation}
\begin{aligned}
  \hat\mu_{\phi,t+k} &= \frac{\mu_\phi(\hat h_{t+k}, \hat C_{t+k})}{\|\mu_\phi(\hat h_{t+k}, \hat C_{t+k})\|_2}, \qquad
  a_{t+k} = \pi_\theta\!\bigl(\hat h_{t+k}, \hat\mu_{\phi,t+k}; x_k^{(0)}\bigr), \\
  \hat s_{t+k+1} &= f_\xi\!\bigl(\hat s_{t+k}, a_{t+k}, c_{t+k}\bigr).
\end{aligned}
\label{eq:imagined-rollout}
\end{equation}
Maximizing the multi-step imagined return directly without auxiliary regularization yields
\begin{equation}
  \max_{\phi,\theta}\; \mathbb E\!\left[
    \sum_{k=0}^{L-1} w_k\,r_{t+k} + w_L V_\omega(\hat s_{t+L}, \hat C_{t+L})
  \right],
  \label{eq:world-model-return}
\end{equation}
where $w_k$ combines discounting and continuation masks, $V_\omega$ is updated with detached return targets, and $r_{t+k}$ evaluates full-body tracking against the complete reference while penalizing action-rate spikes and joint-limit violations (Appendix~\ref{app:rewards}, Tables~\ref{tab:tracking-rewards} and~\ref{tab:control-penalties}). Gradients from the imagined return propagate end-to-end through $f_\xi$, all $K$ Euler steps, and the $\ell_2$-normalized mean $\hat\mu_{\phi,t+k}$ into both the MoE flow decoder ($\theta$) and the sparse prior encoder and mean head ($\phi$), while the unused prior variance head receives zero gradient. Beyond accelerating closed-loop reinforcement learning, this analytical physical feedback directly enhances the physical executability of generative motion predictions while fresh simulator trajectories continuously recalibrate $f_\xi$.


%% file: figure/training_pipeline.tex
\begin{figure}[!t]
\centering
\includegraphics[width=\linewidth]{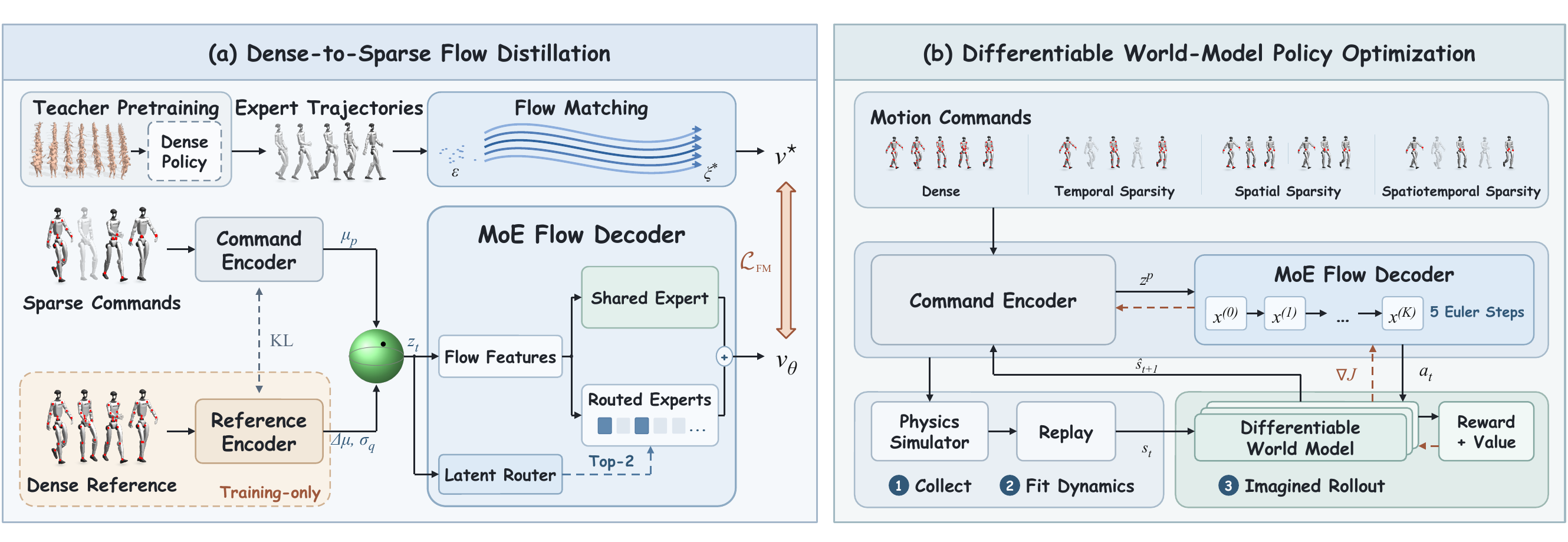}
\caption{\modelname framework. (a) A teacher supplies trajectories for
behavior cloning and labels for DAgger. A sparse command prior learns from a
dense reference posterior, and a latent-routed MoE flow decoder generates
state--action trajectories. (b) Gradients through a differentiable world
model refine the policy in closed loop.}
\label{fig:pipeline}
\end{figure}

%% file: sections/05_experiments.tex
\section{Experiments}
\label{sec:experiments}

In this section, we systematically evaluate \modelname in simulation around three core questions:
\begin{enumerate}[label=(\alph*),leftmargin=*,nosep]
  \item \textbf{General-purpose tracking:} How does \modelname perform as a general-purpose motion tracker compared to state-of-the-art whole-body controllers?
  \item \textbf{Component contributions:} How much do key architectural, training, and observation design choices contribute to overall performance?
  \item \textbf{Spatio-temporal sparsity:} Beyond dense tracking, how accurately and stably does the unified policy handle spatially and temporally sparse commands?
\end{enumerate}

\subsection{Training Data}
\label{sec:training-data}

We construct our training corpus through a three-stage pipeline. First, we aggregate human motion data in SMPL format from publicly available motion-capture and video-reconstructed sources---accounting for over 99\% of our corpus---supplemented by a small set of in-house captures and robot-collected trajectories. Second, we retarget the human motions onto the Unitree G1 morphology using HoloRetarget~\citep{holoretarget}. Third, we apply an in-house post-filtering pipeline to discard clips exhibiting abrupt pose jumps, severe foot sliding, or ground penetration. The resulting corpus contains 1,313,050 clips at 50\,Hz (2,970.80 hours in total), providing broad motion diversity for general-purpose whole-body control. Table~\ref{tab:training-data} lists the retained hours and initial sampling weights per source, which are dynamically updated during training by an adaptive sampler.

\begin{table}[!ht]
\centering
\caption{Training motion sources ordered by initial sampling probability.}
\label{tab:training-data}
\small
\begin{tabular*}{0.88\linewidth}{@{\extracolsep{\fill}}lrr@{}}
\toprule
Motion source & Available hours & Initial sampling (\%) \\
\midrule
MotionMillion \citep{motionmillion} & 1,550.26 & 20 \\
MotionDecode \citep{motiondecode} & 1,103.12 & 20 \\
BONES-SEED \citep{bonesseed} & 196.41 & 15 \\
SnapMoGen \citep{snapmogen} & 50.83 & 15 \\
AMASS \citep{amass} & 47.23 & 7 \\
In-house (PICO) & 4.09 & 5 \\
In-house (Noitom) & 4.04 & 5 \\
LAFAN1 \citep{lafan1} & 4.01 & 5 \\
OMOMO \citep{omomo} & 2.42 & 5 \\
Humanoid Everyday \citep{humanoideveryday} & 8.39 & 3 \\
\midrule
Total & 2,970.80 & 100 \\
\bottomrule
\end{tabular*}
\end{table}
\FloatBarrier

\subsection{Experimental Setup}
\label{sec:experimental_setup}

We evaluate \modelname on two complementary large-scale suites, HumanTracker~\citep{humantracker} and MotionChallenge~\citep{motionchallenge}. Both benchmarks supply dense full-body references (supporting dense and sparse evaluations), standardized protocols with published baselines, and diverse scenarios spanning daily activities, agile movements, and environmental interactions.

\paragraph{HumanTracker (zero-shot / OOD generalization).}
\begin{itemize}[leftmargin=*,nosep]
  \item \textbf{Motion source:} A recently released suite of 2,500 unseen motions across four categories---\emph{Daily}, \emph{Highly Dynamic}, \emph{Interaction}, and \emph{Ground}~\citep{humantracker}. Since these clips are largely absent from the training corpora of competing methods, HumanTracker provides a rigorous testbed for zero-shot and out-of-distribution (OOD) generalization.
  \item \textbf{Evaluation metrics:} We report success rate (SR), joint-angle tracking error (denoted MPJPE in radians, following HumanTracker's convention), and joint jerk.
\end{itemize}

\paragraph{MotionChallenge (large-scale tracking capacity).}
\begin{itemize}[leftmargin=*,nosep]
  \item \textbf{Motion source:} Evaluates all 129,561 motions in the BONES-SEED Unitree G1 29-DoF track~\citep{motionchallenge,bonesseed}. Complementing the unseen-motion evaluation on HumanTracker, MotionChallenge follows the public leaderboard protocol to benchmark large-scale tracking fidelity and capacity across a unified corpus.
  \item \textbf{Evaluation metrics:} We report overall SR and mean per-keypoint position error (MPKPE) under three alignment protocols (XY, XY+Yaw, and World) for both full-rollout and early-termination evaluations.
\end{itemize}
Appendix~\ref{app:benchmark-metrics} details the success criteria, alignment conventions, and metric definitions for both benchmarks.

\paragraph{Implementation and compute.}
A full training run for our standard six-layer-prior policy takes 122.4 hours on eight NVIDIA RTX 5090 GPUs (979 GPU-hours): 19.1 hours for supervised distillation (behavior cloning and four DAgger rounds) and 103.2 hours for 50,000 world-model RL updates (totals summed prior to rounding). For fair comparison, all ablation variants in Section~\ref{sec:ablations} are likewise trained for 50,000 RL updates under identical hardware settings. Full model configurations, data volumes, and compute breakdowns are detailed in Appendices~\ref{app:main-config} and~\ref{app:training-schedule}.

\subsection{Comparison with State-of-the-Art Trackers}
\label{sec:main_results}

\begin{table}[!ht]
\centering
\caption{Quantitative comparison on HumanTracker. Best values per column are in \textbf{bold}.}
\label{tab:humantracker}
\small
\setlength{\tabcolsep}{3.3pt}
\begin{tabularx}{\linewidth}{@{}Lrrrrrrrr@{}}
\toprule
 & \multicolumn{6}{c}{Success rate (\%) $\uparrow$} & & \\
\cmidrule(lr){2-7}
Method & Daily & Dynamic & Interaction & Ground & Avg. & Overall & \shortstack{MPJPE $\downarrow$\\(rad)} & \shortstack{Jerk $\downarrow$\\(rad/s$^3$)} \\
\midrule
TWIST2 \citep{humantracker} & 58.70 & 39.20 & 91.60 & 0.00 & 47.38 & 67.16 & 0.1250 & 824.9 \\
HEFT \citep{heft,humantracker} & 93.20 & 75.70 & 97.50 & 15.90 & 70.58 & 88.13 & 0.0675 & 329.9 \\
GRIT v0.0.1 \citep{grit,humantracker} & 90.20 & 65.70 & 96.50 & 11.00 & 65.85 & 85.13 & 0.1189 & 293.5 \\
ScaleBFM-M \citep{scalebfm,humantracker} & 90.80 & 72.40 & 96.50 & 17.70 & 69.35 & 86.52 & 0.0953 & \textbf{220.1} \\
SONIC \citep{sonic,humantracker} & 93.70 & 83.60 & 97.40 & 19.50 & 73.55 & 89.40 & 0.1231 & 316.2 \\
SONIC 1.1 \citep{sonic,humantracker} & \textbf{94.40} & 84.70 & \textbf{97.80} & 13.40 & 72.58 & 89.52 & 0.1094 & 329.4 \\
HoloMotion v1.4.0 \citep{holomotion1,humantracker} & 84.00 & 63.40 & 94.90 & 18.30 & 65.15 & 82.24 & 0.0962 & 444.7 \\
Our dense teacher & 90.55 & 84.70 & 96.98 & 28.05 & 75.07 & 88.64 & 0.0811 & 366.9 \\
\textbf{\modelname} & 93.22 & \textbf{88.43} & 96.80 & \textbf{28.66} & \textbf{76.78} & \textbf{90.04} & \textbf{0.0580} & 340.5 \\
\bottomrule
\end{tabularx}
\end{table}
\FloatBarrier

\paragraph{HumanTracker.}
On unseen motions (Table~\ref{tab:humantracker}), \modelname is the only method that achieves both the highest success rate and the lowest joint tracking error, whereas prior baselines trade one off against the other. Specifically, \modelname reaches 76.78\% Avg.\ SR, 90.04\% Overall SR, and 0.0580\,rad MPJPE---surpassing SONIC by 3.23\,pp in Avg.\ SR with less than half its joint error (0.1231\,rad), and outperforming HEFT in both Avg.\ SR (+6.20\,pp) and MPJPE (-14.1\%). The largest gains occur on the hardest OOD subsets, \emph{Highly Dynamic} (88.43\%) and \emph{Ground} (28.66\%, +9.16\,pp over SONIC), indicating stronger zero-shot stability under agile and contact-rich dynamics. Its higher joint jerk (340.5 vs.\ 220.1\,rad/s$^3$ for ScaleBFM-M) results from independent per-step sampling in the flow decoder. We exclude Humanoid-GPT since its released evaluator uses privileged ground-truth pelvis tracking errors~\citep{humantracker}.

\begin{table}[!ht]
\centering
\caption{Quantitative comparison on MotionChallenge. Best values per column are in \textbf{bold}.}
\label{tab:motionchallenge}
\small
\setlength{\tabcolsep}{3.8pt}
\begin{tabularx}{\linewidth}{@{}Lrrrrrrr@{}}
\toprule
& \multicolumn{3}{c}{Full-rollout MPKPE (mm) $\downarrow$}
& \multicolumn{3}{c}{Early-termination MPKPE (mm) $\downarrow$} & \\
\cmidrule(lr){2-4}\cmidrule(lr){5-7}
Method & XY & XY+Yaw & World & XY & XY+Yaw & World & Overall SR (\%) $\uparrow$ \\
\midrule
HEFT PMG \citep{heft} & 27.41 & 25.53 & 123.24 & 26.92 & 25.06 & 120.98 & 99.16 \\
MimicLite \citep{mimiclite} & 30.62 & 29.52 & 109.09 & 29.53 & 28.48 & 105.86 & 98.55 \\
ScaleBFM-M \citep{scalebfm} & 30.81 & 31.21 & 183.01 & 30.76 & 31.16 & 182.71 & \textbf{99.80} \\
GRIT \citep{grit} & 32.64 & 32.11 & 164.15 & 32.15 & 31.65 & 162.54 & 99.52 \\
SONIC Original \citep{sonic} & 33.48 & 32.51 & 111.65 & 32.93 & 31.96 & 110.50 & 99.75 \\
SONIC 1.1 \citep{sonic,motionchallenge} & 35.95 & 32.81 & 160.14 & 35.14 & 32.01 & 157.87 & 99.48 \\
HoloMotion v1.4.0 \citep{holomotion1,motionchallenge} & 43.22 & 37.30 & 173.10 & 40.67 & 34.80 & 168.97 & 98.08 \\
\textbf{\modelname} & \textbf{19.66} & \textbf{19.08} & \textbf{104.57} & \textbf{19.64} & \textbf{19.07} & \textbf{104.11} & 99.27 \\
\bottomrule
\end{tabularx}
\end{table}

\paragraph{MotionChallenge.}
Across all 129,561 motions (Table~\ref{tab:motionchallenge}), \modelname achieves the lowest tracking error on every local and global MPKPE metric under both evaluation protocols. It reduces full-rollout XY and XY+Yaw errors to 19.66\,mm and 19.08\,mm (28.3\% and 25.3\% below HEFT PMG) while attaining the lowest world-frame error (104.57\,mm vs.\ 109.09\,mm for MimicLite). By contrast, baselines with slightly higher completion rates suffer substantially larger global drift or local error (e.g., 183.01\,mm world MPKPE for ScaleBFM-M at 99.80\% SR; 33.48\,mm XY error for SONIC at 99.75\% SR). \modelname maintains 99.27\% overall SR, and its nearly identical full-rollout and early-termination errors indicate consistent tracking across the corpus. We analyze the remaining trade-off between global drift suppression and local completion rate under proprioception-only control in Section~\ref{sec:ablations}.

Overall, closed-loop reinforcement learning enables \modelname to outperform existing trackers on dense tracking benchmarks in both OOD generalization and large-scale tracking accuracy, while supporting spatially and temporally sparse commands within the same unified controller.

\subsection{Ablation Studies}
\label{sec:ablations}

We conduct ablations on HumanTracker to isolate the contributions of each training stage, sparse prior depth, RL training data size, flow integration steps, and RL training target origin.

\begin{figure}[!ht]
\centering
\includegraphics[width=0.98\linewidth]{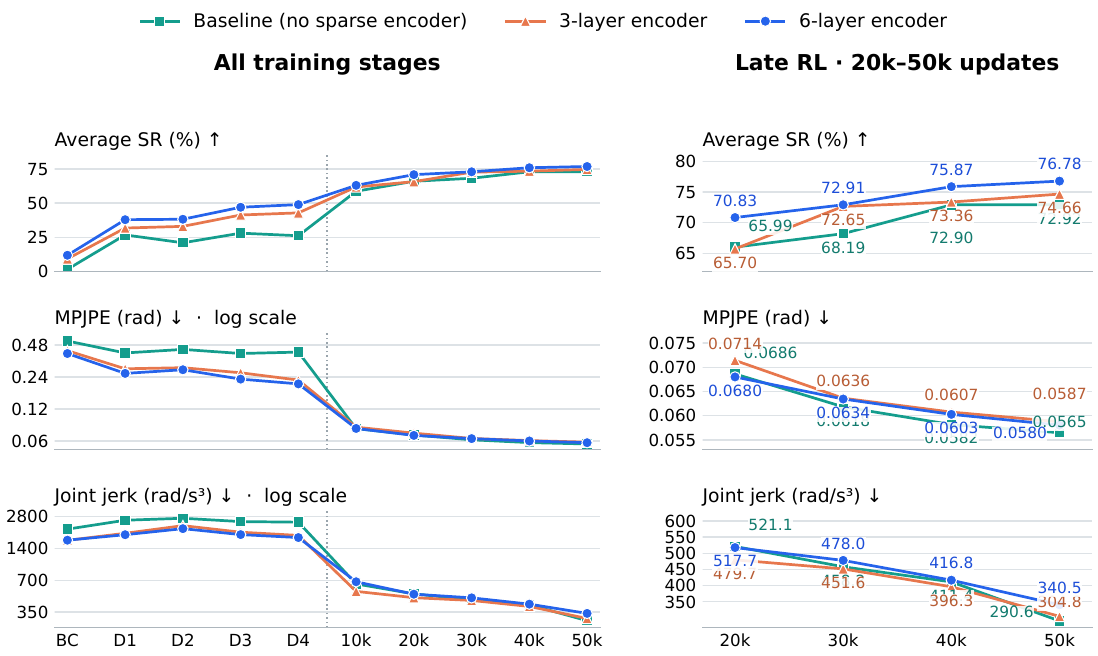}
\caption{HumanTracker performance across training stages. Left panels show behavior cloning (BC), four DAgger rounds (D1--D4), and world-model RL (with logarithmic scales for MPJPE and joint jerk). Right panels magnify the 20k--50k RL regime on linear axes with exact values labeled.}
\label{fig:training-stage-ablation}
\end{figure}

\paragraph{Performance across training stages.}
Figure~\ref{fig:training-stage-ablation} tracks the baseline (no sparse prior encoder) alongside the 3-layer and 6-layer encoder variants across behavior cloning, four rounds of DAgger, and world-model RL. After DAgger alone, all three models still exhibit elevated joint-angle error and jerk, whereas world-model RL sharply improves tracking accuracy and smoothness across every configuration. For the 6-layer encoder, 50,000 RL updates lift average SR from 48.94\% to 76.78\%, cut MPJPE from 0.2073 to 0.0580\,rad, and reduce mean joint jerk from 1,777 to 340.5\,rad/s$^3$. Backpropagating multi-step physical rewards through the differentiable world model and iterative flow solver directly aligns action synthesis with closed-loop dynamics, overcoming the compounding errors of open-loop imitation.

\begin{figure}[!ht]
\centering
\begin{minipage}[c]{0.57\linewidth}
\centering
\includegraphics[width=\linewidth]{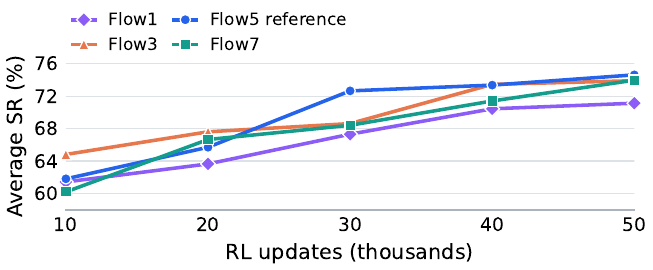}
\end{minipage}\hfill
\begin{minipage}[c]{0.39\linewidth}
\centering\small
\textbf{Policy inference latency}\\[0.6em]
\setlength{\tabcolsep}{3pt}
\begin{tabular}{@{}rrrr@{}}
\toprule
\shortstack{Encoder\\layers} & \shortstack{Flow\\steps} & \shortstack{Mean\\(ms)} & \shortstack{P99\\(ms)} \\
\midrule
0 & 5 & 4.1 & 5.6 \\
3 & 5 & 5.2 & 6.7 \\
6 & 5 & 6.4 & 7.8 \\
\midrule
3 & 1 & 2.0 & 3.5 \\
3 & 3 & 3.6 & 5.1 \\
3 & 7 & 6.9 & 8.3 \\
\bottomrule
\end{tabular}
\end{minipage}
\caption{HumanTracker average SR across flow integration steps (left) and policy inference latency on Jetson Orin NX (right). Timings exclude retargeting, forward kinematics, and observation construction.}
\label{fig:flow-step-ablation}
\end{figure}

\FloatBarrier
\paragraph{Scaling with model size.}
Adding a transformer sparse prior consistently boosts average SR, scaling monotonically with encoder depth (Figure~\ref{fig:training-stage-ablation}). After four DAgger rounds, average SR rises from 26.19\% (0 layers) to 42.89\% (3 layers) and 48.94\% (6 layers); after 50,000 RL updates, this hierarchy persists at 72.92\%, 74.66\%, and 76.78\%, respectively. As shown on the right of Figure~\ref{fig:flow-step-ablation}, this accuracy gain incurs only modest onboard compute cost on a Jetson Orin NX: at $K=5$ flow steps, mean inference latency scales roughly linearly from 4.1\,ms (0 layers) to 5.2\,ms (3 layers) and 6.4\,ms (6 layers).

\begin{figure}[!ht]
\centering
\includegraphics[width=0.98\linewidth]{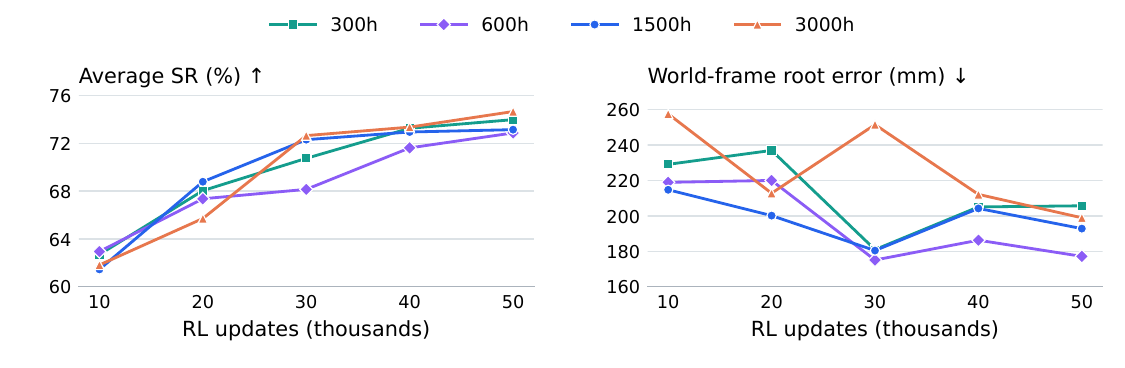}
\caption{Effect of RL training data size on HumanTracker. Left: category-average SR. Right: unaligned world-frame root error.}
\label{fig:data-size-scaling}
\end{figure}

\paragraph{Scaling with data size.}
Figure~\ref{fig:data-size-scaling} ablates the RL training data size across 300\,h, 600\,h, 1,500\,h, and the full $\sim$3,000\,h corpus (2,970.80\,h) using the 3-layer encoder over 50,000 updates. All four scales achieve comparable world-frame root error, while the full corpus reaches the highest final average SR (74.66\%). Notably, even at 600\,h---matching the approximate data scale of ScaleBFM~\citep{scalebfm}---or 300\,h, our 3-layer policy attains roughly 73\%--74\% average SR, still outperforming ScaleBFM-M (69.35\%, Table~\ref{tab:humantracker}) by over 3\,pp. This confirms that our gains stem primarily from the generative flow policy and world-model RL rather than data volume alone, while larger corpora further improve OOD generalization.

\paragraph{Number of flow integration steps.}
Figure~\ref{fig:flow-step-ablation} compares Euler integration steps $K \in \{1, 3, 5, 7\}$ for 3-layer encoder policies initialized from the same DAgger-4 checkpoint. At 50,000 RL updates, average SR improves from 71.13\% ($K=1$) to 73.90\% ($K=3$) and peaks at 74.66\% ($K=5$). Increasing to $K=7$ yields 73.96\%, as GPU memory limits required halving the per-GPU actor batch size from 1,024 to 512 during backpropagation through the solver; thus, $K=5$ offers the best accuracy--compute--memory trade-off. Across all evaluated encoder depths and flow-step counts, both mean and P99 latencies on Jetson Orin NX remain under 9\,ms---comfortably within the 20\,ms (50\,Hz) control budget---with no observable latency-induced jitter on hardware.

\FloatBarrier

\begin{figure}[!ht]
\centering
\includegraphics[width=0.98\linewidth]{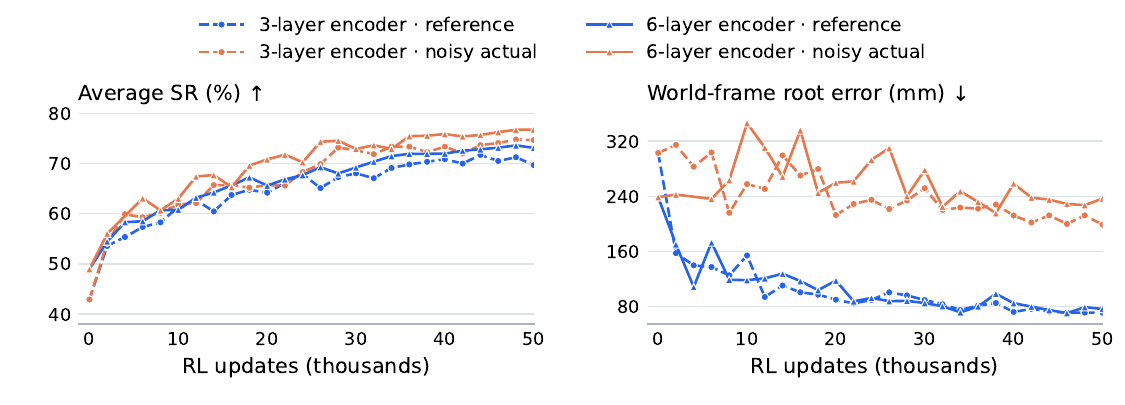}
\caption{Effect of the RL training pelvis origin on HumanTracker. Left: category-average SR. Right: unaligned world-frame root error.}
\label{fig:pelvis-origin-ablation}
\end{figure}

\paragraph{Global versus local tracking performance.}
To examine the trade-off between local balance robustness and long-horizon global drift observed in Section~\ref{sec:main_results}, we ablate the pelvis origin used to anchor target positions during RL. While proprioception-only deployment lacks external localization and must use the reference pelvis origin, RL training can anchor target positions to either (1) the robot's actual pelvis position perturbed by $\pm 5$\,cm per-axis uniform noise, or (2) the reference pelvis origin, holding DAgger initialization, robot orientation, and reference-origin evaluation fixed within each encoder depth. Figure~\ref{fig:pelvis-origin-ablation} confirms this trade-off across both depths: noisy actual-origin training yields higher average SR but larger world-frame root error, whereas reference-origin training exhibits the opposite trend. Matching training to the reference origin enforces accurate open-loop displacement under world-frame penalties, sharply reducing global drift at a modest cost to SR; conversely, noisy actual-origin training acts as target perturbation that bolsters local balance, but accumulates greater drift once global feedback is absent at test time.

\FloatBarrier

\subsection{Spatio-Temporal Sparse-Command Evaluation}
\label{sec:sparse-eval}

Beyond dense full-body tracking, we evaluate \modelname under spatial sparsity (five-point tracking and single-point velocity control) and temporal sparsity (fixed and mixed waypoint intervals).

\paragraph{Spatial sparsity: five-point tracking.}
We compare \modelname against ScaleBFM-M~\citep{scalebfm} on HumanTracker under identical reference clips and termination criteria (Table~\ref{tab:sparse-humantracker}). \modelname achieves 75.92\% category-average SR---outperforming ScaleBFM-M (69.29\%) by 6.63\,pp---and reduces MPJPE from 0.1399 to 0.1254\,rad. Gains are most pronounced on Highly Dynamic (85.45\% vs.\ 71.27\%) and Ground (31.10\% vs.\ 17.68\%) motions, demonstrating strong full-body completion under heavily masked spatial commands.

\begin{table}[!ht]
\centering
\caption{Five-point spatial sparsity comparison on HumanTracker against ScaleBFM. Best values per column are in \textbf{bold}.}
\label{tab:sparse-humantracker}
\small
\setlength{\tabcolsep}{3.5pt}
\begin{tabularx}{\linewidth}{@{}Lrrrrrrrr@{}}
\toprule
 & \multicolumn{6}{c}{Success rate (\%) $\uparrow$} & & \\
\cmidrule(lr){2-7}
Method & Daily & Dynamic & Interaction & Ground & Avg. & Overall & \shortstack{MPJPE $\downarrow$\\(rad)} & \shortstack{Jerk $\downarrow$\\(rad/s$^3$)} \\
\midrule
ScaleBFM-M \citep{scalebfm} & 90.66 & 71.27 & \textbf{97.53} & 17.68 & 69.29 & 86.80 & 0.1399 & \textbf{258.2} \\
\textbf{\modelname} & \textbf{90.97} & \textbf{85.45} & 96.16 & \textbf{31.10} & \textbf{75.92} & \textbf{88.72} & \textbf{0.1254} & 339.5 \\
\ifshowthreepoint
\midrule
\multicolumn{9}{@{}l}{\textit{3-point sparse control}} \\
ScaleBFM-M \citep{scalebfm} & 68.28 & 44.78 & 93.88 & 17.07 & 56.00 & 73.60 & 0.1583 & 284.3 \\
\textbf{\modelname} & 57.29 & 44.40 & 89.95 & 12.20 & 50.96 & 67.24 & 0.1608 & 316.1 \\
\fi
\bottomrule
\end{tabularx}
\end{table}

\paragraph{Spatial sparsity: single-point velocity control.}
When only the root is commanded, limb coordination and gait synthesis are completely unspecified. Whereas tracking-only frameworks such as SONIC~\citep{sonic} rely on an upstream motion generator to synthesize full-body references from velocity commands, \modelname natively generates whole-body locomotion directly from single-point targets. Following SONIC's velocity-tracking protocol, we evaluate \modelname against the public OpenHomie release~\citep{homie} across 200 five-second trials (four directions $\times$ 50 commanded speeds from 0.1 to 5.0\,m/s), with each system retaining its native robot dynamics and policy interface.

\begin{figure}[!ht]
\centering
\includegraphics[width=\linewidth]{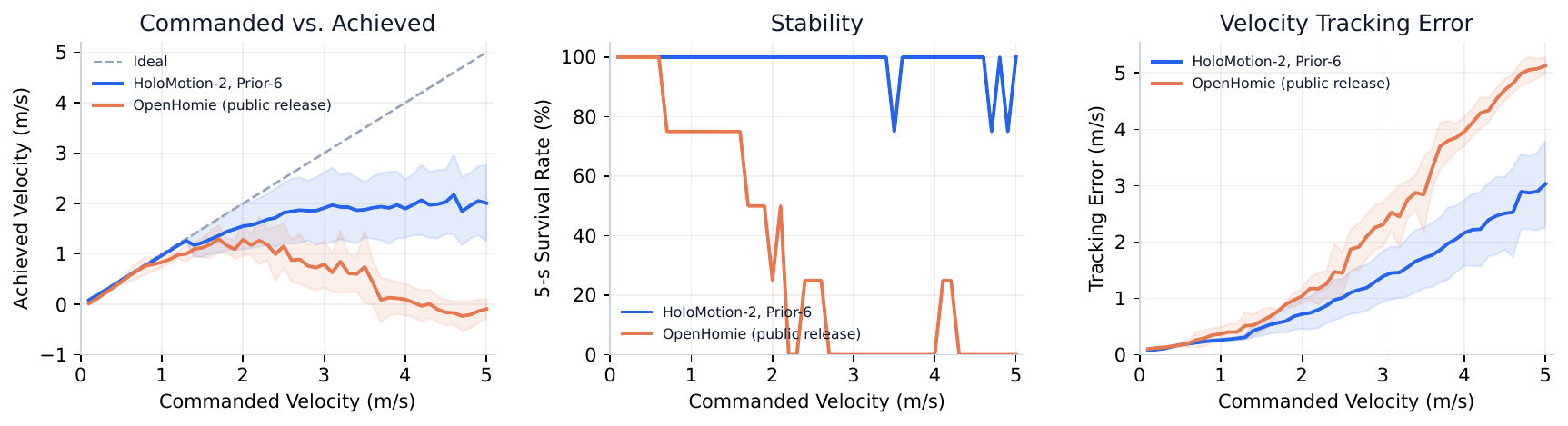}
\caption{Single-point velocity tracking against OpenHomie across 200 five-second trials (four directions, 0.1--5.0\,m/s), following the layout of SONIC~\citep{sonic}. Left: commanded versus achieved velocity. Middle: 5-s survival rate. Right: mean 2-D velocity error. Curves average four directions with $\pm 1$ standard deviation shading.}
\label{fig:sparse-velocity}
\end{figure}

As shown in Figure~\ref{fig:sparse-velocity}, \modelname achieves 98.5\% five-second survival with a mean velocity error of 1.217\,m/s, compared with 34.0\% and 2.100\,m/s for OpenHomie. Under high speed commands, achieved velocity saturates near 2\,m/s due to physical locomotion limits, so survival primarily reflects balance robustness rather than exact high-speed attainment.
\FloatBarrier

\paragraph{Temporal sparsity.}
High-level planners and VLA models typically issue waypoints well below 50\,Hz and at irregular intervals. We evaluate temporal sparsity on HumanTracker over waypoint intervals $\Delta t \in \{0.2, 0.4, 0.6, 0.8, 1.0\}$\,s under two schedules:
\begin{itemize}[leftmargin=*,nosep]
  \item \textbf{Fixed-interval updates:} Each rollout maintains a constant interval $\Delta t$ (reported as the equal-weight mean across the five intervals), reflecting periodic 1--5\,Hz planners.
  \item \textbf{Mixed-interval updates:} After each waypoint, the next interval is drawn uniformly from the five candidates, mirroring asynchronous planners with variable inference latency or event-driven replanning.
\end{itemize}
Throughout each rollout, we report frame-level local and global keypoint errors (Local, Global), planar root position error (XY), and root yaw error (Yaw) against the dense reference, together with command-level arrival rate (AR); full definitions and interface details are provided in Appendix~\ref{app:benchmark-metrics}.

\begin{table}[!ht]
\centering
\caption{Temporal sparsity comparison on HumanTracker. Best values per column are in \textbf{bold}.}
\label{tab:temporal-sparsity}
\small
\setlength{\tabcolsep}{2.8pt}
\begin{tabularx}{\linewidth}{@{}L*{10}{r}@{}}
\toprule
 & \multicolumn{5}{c}{Fixed} & \multicolumn{5}{c}{Mixed} \\
\cmidrule(lr){2-6}\cmidrule(l){7-11}
Method & \shortstack{Local $\downarrow$\\(cm)} & \shortstack{Global $\downarrow$\\(cm)} & \shortstack{XY $\downarrow$\\(m)} & \shortstack{Yaw $\downarrow$\\($^{\circ}$)} & \shortstack{AR $\uparrow$\\(\%)} & \shortstack{Local $\downarrow$\\(cm)} & \shortstack{Global $\downarrow$\\(cm)} & \shortstack{XY $\downarrow$\\(m)} & \shortstack{Yaw $\downarrow$\\($^{\circ}$)} & \shortstack{AR $\uparrow$\\(\%)} \\
\midrule
METATRON G1 \citep{metatron} & 5.63 & 12.87 & 0.1060 & 10.69 & 96.30 & 5.70 & 12.00 & 0.0959 & 10.31 & 97.14 \\
\mbox{\textbf{\modelname (5-point)}} & 3.23 & 6.12 & 0.0456 & 3.70 & 99.65 & 3.49 & 6.28 & 0.0454 & 4.17 & 99.64 \\
\mbox{\textbf{\modelname (14-point)}} & \textbf{2.79} & \textbf{5.83} & \textbf{0.0444} & \textbf{3.61} & \textbf{99.74} & \textbf{3.09} & \textbf{6.05} & \textbf{0.0447} & \textbf{4.11} & \textbf{99.72} \\
\bottomrule
\end{tabularx}
\end{table}

We benchmark against METATRON G1~\citep{metatron} using 5-point waypoints to match its command interface, and additionally evaluate 14-point waypoints on the same \modelname checkpoint (Table~\ref{tab:temporal-sparsity}). Across both Fixed and Mixed schedules, \modelname (5-point) outperforms METATRON G1 on every metric: it roughly halves global tracking error (6.12 vs.\ 12.87\,cm in Fixed; 6.28 vs.\ 12.00\,cm in Mixed), cuts Mixed local error from 5.70 to 3.49\,cm, and raises Mixed AR from 97.14\% to 99.64\%. Providing 14-point waypoints to the same checkpoint further lowers Mixed local and global errors to 3.09\,cm and 6.05\,cm, reaching 99.72\% AR.

%% file: sections/05_application.tex
\section{Applications}
\label{sec:application}

While Section~\ref{sec:experiments} benchmarks \modelname in simulation, physical deployment introduces unmodeled contact dynamics, actuator limits, and sensor noise. To validate sim-to-real transfer under these conditions, we evaluate \modelname on a Unitree G1 across diverse real-world downstream applications. Across loco-manipulation and interactive boxing, the exact same unified policy checkpoint is deployed without task-specific retraining, whether the robot operates unloaded or carries a payload backpack, driven by upstream planners through dense or spatio-temporally sparse commands. We further demonstrate the extensibility of our architecture on table tennis by incorporating an auxiliary task-specific command token.

\subsection{Loco-Manipulation}
\label{sec:app-loco-manip}

We integrate \modelname with HoloBrain-0.5~\citep{holobrain05} in a unified data-to-deployment stack for autonomous loco-manipulation. During data collection, \modelname serves as the whole-body teleoperation controller to gather physically feasible demonstrations for training HoloBrain-0.5. At deployment time, HoloBrain-0.5 predicts task-driven motion commands from onboard perception, and \modelname executes them in closed loop at 50\,Hz. To prioritize contact precision during manipulation, the high-level policy interfaces with \modelname via dense reference-point commands.

\begin{figure}[!p]
\centering
\includegraphics[width=\linewidth]{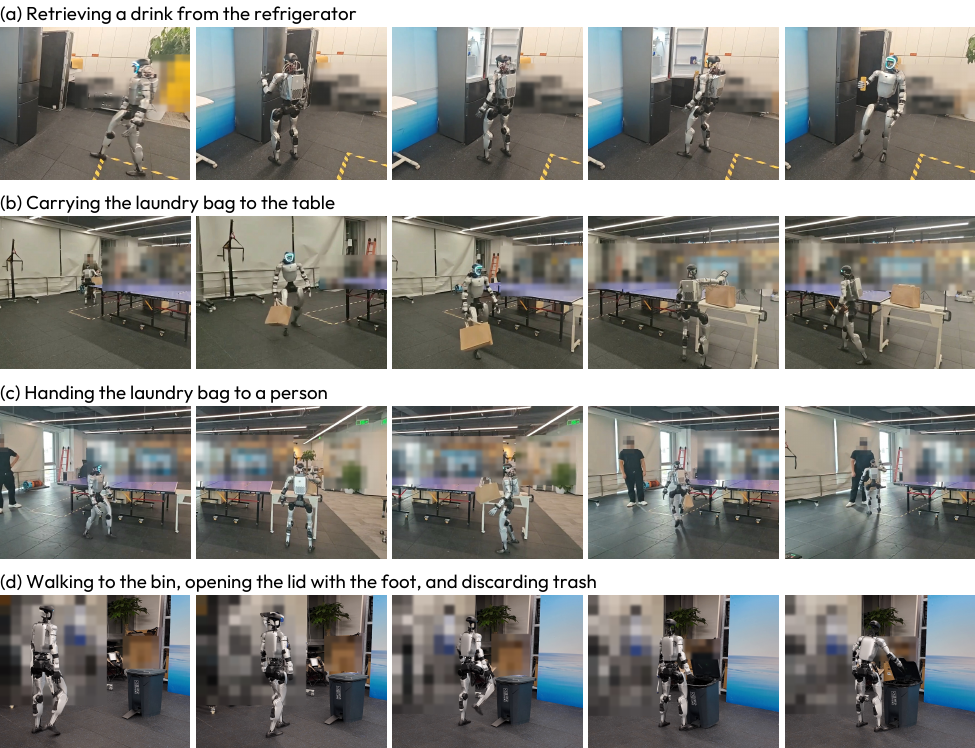}
\caption{Real-world loco-manipulation on the Unitree G1 executed by \modelname. (a)--(c) Autonomous HoloBrain-0.5 rollouts for refrigerator retrieval and clothes delivery; (d) teleoperated pedal-bin trash disposal.}
\label{fig:app-loco-manip}
\end{figure}

We quantitatively evaluate this autonomous stack across two multi-stage household tasks:
\begin{itemize}[leftmargin=*,nosep]
  \item \textbf{Refrigerator water retrieval:} Approach the refrigerator, pull open the hinged door, grasp a water bottle, and step back smoothly.
  \item \textbf{Clothes fetching and delivery:} Grasp garments bimanually, turn in place, and walk to deliver and hand them off at a target location.
\end{itemize}
Figure~\ref{fig:app-loco-manip}(a)--(c) illustrates real-world rollouts of these tasks, and Table~\ref{tab:app-loco-manip} reports the demonstration volume, trial count, and real-world success rate for each task, demonstrating reliable whole-body execution across contact-rich and payload-carrying transitions.

Beyond autonomous execution, we further verify \modelname's whole-body loco-manipulation capability under teleoperation on a coordinated \emph{pedal-bin trash disposal} task (Figure~\ref{fig:app-loco-manip}(d)). The robot walks up to a lidded bin, shifts its weight onto one leg to depress the foot pedal, and reaches over the open lid to discard trash---demonstrating simultaneous lower-body contact actuation and upper-body balance control.

\begin{table}[!ht]
\centering
\caption{Real-world autonomous loco-manipulation evaluation on the Unitree G1 using HoloBrain-0.5 with dense reference-point commands executed by \modelname.}
\label{tab:app-loco-manip}
\small
\begin{tabularx}{\linewidth}{@{}Lrrr@{}}
\toprule
Task & Training data (demos) & Real-world trials & Success rate (\%) \\
\midrule
Refrigerator water retrieval & 700 & 20 & 80.0 \\
Clothes fetching and delivery & 1300 & 20 & 85.0 \\
\bottomrule
\end{tabularx}
\end{table}

\subsection{Boxing}
\label{sec:app-boxing}

\begin{figure}[!p]
\centering
\IfFileExists{figure/app_boxing.pdf}{\includegraphics[width=\linewidth]{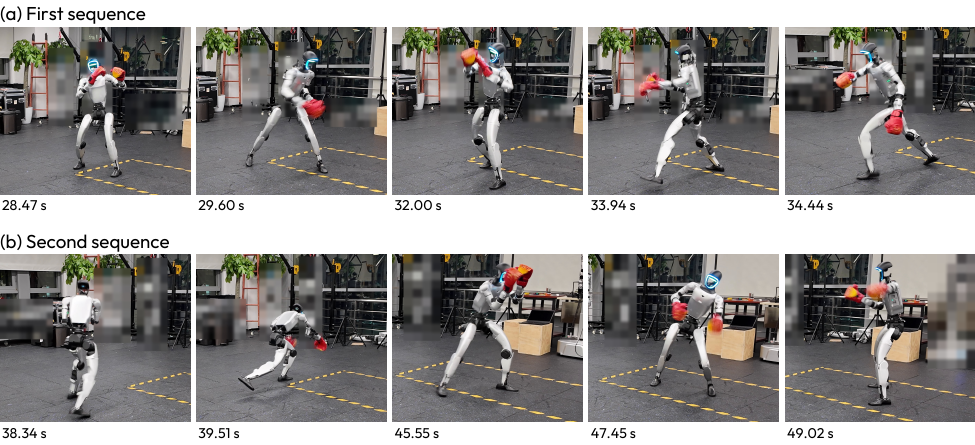}}{%
\fbox{\parbox[c][3\baselineskip][c]{0.94\linewidth}{\centering Boxing rollout keyframes pending.}}}
\caption{Interactive boxing on the Unitree G1. Driven by an upstream planner such as HoloBoxing~\citep{holoboxing}, \modelname executes rapid punches, dodges, and footwork using five-point sparse commands (pelvis, wrists, ankles).}
\label{fig:app-boxing}
\end{figure}

Interactive boxing stresses high-dynamic tracking and balance recovery under explosive punches, rapid dodges, and abrupt momentum shifts. We couple \modelname with an upstream sparring planner such as HoloBoxing~\citep{holoboxing}, which generates interactive combat trajectories online in response to opponent motions. Unlike the dense interface used in loco-manipulation, this deployment operates purely under \textbf{five-point sparse control}: the planner commands only the pelvis, both wrists, and both ankles, delegating torso rotation, arm-joint coordination, and lower-body stabilization entirely to \modelname. Despite high velocities and rapid direction reversals, the same unified policy tracks planned strikes and evasive footwork smoothly across consecutive exchanges (Figure~\ref{fig:app-boxing}).

\subsection{Table Tennis}
\label{sec:app-table-tennis}

We further show that \modelname's architecture extends naturally beyond pure pose tracking to heterogeneous task commands. Specifically, we inject the predicted 3D ball-striking position as an additional command token alongside the sparse reference pose inputs, enabling the transformer prior to jointly encode kinematic constraints and the interception target without altering the core decoder.

Under this extended command interface, the policy achieves a \textbf{62.5}\% return success rate in simulation and sustains rallies of up to \textbf{18} consecutive returns in real-world play. These results confirm that our framework generalizes cleanly to task-centric auxiliary commands, establishing a foundation for broader multimodal control in future work.
\FloatBarrier

%% file: sections/06_conclusion.tex
\section{Conclusion}
\label{sec:conclusion}

We introduced \modelname, a generative behavior foundation model for humanoid whole-body control that unifies a timed sparse command interface, a latent-routed MoE state--action flow decoder, and closed-loop policy optimization through a differentiable world model. Supervised dense-to-sparse distillation via behavior cloning and DAgger provides a strong generative initialization, while world-model reinforcement learning resolves compounding closed-loop errors---raising HumanTracker mean success rate from 48.94\% to 76.78\%, reducing MPJPE from 0.2073 to 0.0580\,rad, and surpassing published baselines on HumanTracker as well as all local and global tracking error metrics on MotionChallenge.

%% file: sections/07_future_work.tex
\section{Limitations and Future Work}
\label{sec:future_work}

While \modelname demonstrates strong whole-body tracking fidelity and physical robustness across diverse tasks, several key directions remain on the path toward general-purpose humanoid intelligence.

\paragraph{Toward Any-Command Conditioning.}
Although \modelname unifies dense references with spatially and temporally sparse pose constraints, it has not yet achieved universal ``any-command'' control across arbitrary task specifications. Crucially, our tokenized sparse prior and latent-routed MoE flow decoder already establish a unified training framework capable of absorbing heterogeneous conditioning signals. Extending this architecture to natively support broader multimodal and increasingly sparse command interfaces is a primary focus of our ongoing work.

\paragraph{Contact-Aware Whole-Body Interaction.}
We aim to extend \modelname to physical interactions involving objects and the environment. By incorporating object states and contact objectives into our sparse command interface---and expanding the differentiable world model to coupled robot--object dynamics---we will explore coordinated pushing, carrying, and environmental support under changing contacts.

\paragraph{Exploring More Efficient System Interfaces.}
Previously, limited by the generality of low-level controllers (system~0) and their reliance on high-frequency control commands, substantial constraints were imposed on the output space of high-level embodied models (system~2\&system~1). Looking ahead, we aim to co-design more efficient, tightly coupled interfaces---connecting \modelname directly with high-level embodied agents to unlock true whole-body intelligence in unstructured real-world environments.

%% file: sections/08_appendix.tex
\clearpage
\appendix
\section{Implementation and Reproducibility}
\label{sec:appendix}

\subsection{Main Configuration}
\label{app:main-config}

Our default policy uses a six-layer sparse encoder and five Euler integration steps for flow generation. Table~\ref{tab:config} summarizes its architecture, control frequencies, and temporal command sampling. Detailed input and output representations are provided in Appendix~\ref{app:policy-io}.

\begin{table}[!ht]
\centering
\caption{Main model and environment hyperparameters.}
\label{tab:config}
\small
\begin{tabularx}{\linewidth}{@{}lL@{}}
\toprule
Component & Setting \\
\midrule
Robot & Unitree G1, 29 actuated joints \\
Control frequencies & 50\,Hz policy, 200\,Hz physics \\
Sparse encoder & 6 layers, width 512, 4 attention heads, FFN width 2,048, 64-dimensional latent \\
Flow decoder & Width 512, 1 shared expert, 2 of 1,024 routed experts, 5 Euler integration steps \\
Spatial command sampling & 14-, 5-, 3-, and 1-point commands with sampling probabilities of 33\%, 33\%, 29\%, and 5\%, respectively. \\
Temporal command sampling & 5 candidate target slots; dense 20\%, intermediate 60\%, sparse 20\% \\
\bottomrule
\end{tabularx}
\end{table}

\subsection{Domain Randomization}
\label{app:domain-randomization}

To bridge the sim-to-real gap and improve robustness against dynamics modeling discrepancies, we apply simulation domain randomization to contact dynamics, actuator properties, and external disturbances during flat-ground training, as summarized in Table~\ref{tab:dr-physics}. Sliding friction (shared across both feet with higher contact priority than the ground), torso center of mass (COM), neutral joint offsets, and $K_p$/$K_d$ gains are sampled per environment at initialization, while actuator command delay is resampled at episode reset. Intermittent base-velocity disturbances split world-model replay sequences without triggering episode termination, and the realized physical parameters condition only the world model while remaining hidden from the actor.

\begin{table}[!ht]
\centering
\caption{Simulation domain randomization in flat-ground training. $\mathcal{U}[a,b]$ and $\mathcal{DU}[a,b]$ denote continuous and discrete uniform distributions, respectively; all dimensional quantities use SI units (m, rad, s) unless noted.}
\label{tab:dr-physics}
\small
\begin{tabularx}{\linewidth}{@{}lL@{}}
\toprule
Parameter & Distribution / Range \\
\midrule
Foot sliding friction (shared) & Uniform sampling over 64 discrete friction levels in $[0.3,\, 1.2]$ \\
Torso COM offset & $x:\,\mathcal{U}[-0.075,\, 0.075]$;\; $y,z:\,\mathcal{U}[-0.10,\, 0.10]$ \\
Neutral joint-angle offset & $\mathcal{U}[-0.01,\, 0.01]$ \\
PD gain scaling ($K_p$, $K_d$) & $\mathcal{U}[0.9,\, 1.1]$ (per actuator) \\
Actuator command delay & $\mathcal{DU}[0,\, 2]$ substeps (at 200\,Hz) \\
Base-velocity disturbance & Interval $\mathcal{U}[1,\, 3]$;\; $v_{x,y}\!:\pm 0.50$, $v_z\!:\pm 0.20$;\; $\omega_{x,y}\!:\pm 0.52$, $\omega_z\!:\pm 0.78$ \\

\bottomrule
\end{tabularx}
\end{table}

To further improve robustness to real-world sensing and state-estimation errors, observation and sparse-goal perturbations are injected into the policy inputs during both physical and imagined rollouts, as listed in Table~\ref{tab:dr-noise}. We distinguish per-component goal-feature noise from pelvis localization noise: the latter adds a shared offset $\boldsymbol{\varepsilon}_{\mathrm{xyz}} \sim \mathcal{U}[-a, a]^3$ to the measured pelvis position before transforming all targets into the pelvis frame, $\mathbf{p}_{\mathrm{local}} = \mathbf{R}_{\mathrm{pelvis}}^{\top}(\mathbf{p}_{\mathrm{goal}} - (\mathbf{p}_{\mathrm{pelvis}} + \boldsymbol{\varepsilon}_{\mathrm{xyz}}))$, without altering the simulated state. Alternative pelvis-origin conventions (reference pelvis vs.\ actual pelvis with $\pm 5$\,cm localization noise) are evaluated as controlled variants in Section~\ref{sec:ablations}.

\begin{table}[!ht]
\centering
\caption{Observation and sparse-goal perturbations applied to policy inputs during both physical and imagined world-model rollouts (SI units: m, rad, s).}
\label{tab:dr-noise}
\small
\begin{tabularx}{\linewidth}{@{}l l L@{}}
\toprule
Input Feature & Distribution / Range & Injection Point \\
\midrule
\multicolumn{3}{@{}l}{\textit{Proprioceptive observation noise}} \\
Projected gravity & $\mathcal{U}[-0.10,\, 0.10]$ & Per component on actor proprioception \\
Base angular velocity (local) & $\mathcal{U}[-0.20,\, 0.20]$ & Per component on actor proprioception \\
Joint position (rel.\ to neutral) & $\mathcal{U}[-0.01,\, 0.01]$ & Per joint on actor proprioception \\
Joint velocity & $\mathcal{U}[-0.50,\, 0.50]$ & Per joint on actor proprioception \\
\midrule
\multicolumn{3}{@{}l}{\textit{Sparse-goal perturbations}} \\
Pelvis XYZ localization noise & $\mathcal{U}[-0.05,\, 0.05]$ (per axis) & Shared offset on measured $\mathbf{p}_{\mathrm{pelvis}}$ before target transform \\
Goal link position & $\mathcal{U}[-0.05,\, 0.05]$ & Per component after local-frame transform (active targets only) \\
Goal link 6D rotation & $\mathcal{U}[-0.05,\, 0.05]$ & Per 6D matrix component, not angle (active targets only) \\
\bottomrule
\end{tabularx}
\end{table}

\FloatBarrier

\subsection{Benchmark Metrics}
\label{app:benchmark-metrics}

\paragraph[HumanTracker]{HumanTracker~\cite{humantracker}.} \href{https://github.com/GalaxyGeneralRobotics/HumanTracker#zero-shot-results}{Official leaderboard} and \href{https://github.com/GalaxyGeneralRobotics/HumanTracker/tree/492f128dc903d1c859d2ff535c549c149f9adda8/src/humantracker/eval/core}{evaluation code}.
\begin{itemize}[leftmargin=*,nosep]
    \item \textbf{SR (\%):} Percentage of clips that reach the end of the reference motion. The whole-body evaluator stops scoring at the first nonfinite generalized position or velocity, a pelvis rotation error above 1\,rad, or an absolute $z$ error above 0.25\,m at the pelvis, either wrist, or either ankle. It measures the pose errors in each tracked body's local axes after removing horizontal pelvis translation and heading.
    \item \textbf{Avg.\ SR / Overall SR (\%):} We report the unweighted mean of the four category success rates and the percentage of all 2,500 clips that succeed, respectively. For external baselines in Table~\ref{tab:humantracker}, we estimate overall SR, MPJPE, and jerk by weighting the published category values by the number of clips in each category.
    \item \textbf{MPJPE (rad):} Mean absolute difference between the executed and reference joint angles, excluding the floating base.
    \item \textbf{Root position error (mm) / Joint jerk (rad/s$^3$):} For the ablations, root error is the mean absolute difference in world-frame pelvis coordinates. Joint jerk is the mean absolute third time difference of the executed joint angles, divided by the control interval cubed.
\end{itemize}

\paragraph[MotionChallenge]{MotionChallenge~\cite{motionchallenge}.} \href{https://github.com/MotionChallenge-Project/MotionChallenge-Core/blob/e4b5c723dcc96be7ede18eb1277ef6ec447e2df0/docs/PROTOCOL.md}{Official protocol}.
\begin{itemize}[leftmargin=*,nosep]
    \item \textbf{Overall SR (\%):} Percentage of clips reaching the reference endpoint without a failure. A clip fails when the absolute pelvis-height error exceeds 0.25\,m or when the actual and reference unit-gravity vectors, each expressed in its own pelvis frame, differ by more than 0.8 in their $z$ component.
    \item \textbf{MPKPE (mm):} Mean 3D Euclidean distance between corresponding robot and reference keypoints. The evaluator averages over 14 body origins and scored frames within each clip, then gives each clip equal weight.
    \item \textbf{XY / XY+Yaw / World:} At each frame, XY centers each pose on its own pelvis in the horizontal plane. XY+Yaw additionally rotates the robot pose about the world vertical to align pelvis heading. World uses the original coordinates. All three retain absolute height.
    \item \textbf{Full rollout / Early termination:} Both scores include failed clips. The full score uses frames through the reference endpoint, including those after failure. The early score uses frames through the first failure frame, or the entire motion if it succeeds.
\end{itemize}

\paragraph[Temporal Sparsity]{Temporal Sparsity.}
Evaluated across all 2,500 HumanTracker motions~\cite{humantracker} in 50\,Hz Unitree G1 simulation using shared waypoint schedules ($\Delta t \in \{0.2, 0.4, 0.6, 0.8, 1.0\}$\,s). Rollouts start from the first reference pose with zero velocity and run without early termination; policies observe only the active waypoint, switching at each deadline and holding the final waypoint until clip end.
\begin{itemize}[leftmargin=*,nosep]
    \item \textbf{Local / Global (cm):} Frame-level mean 3D Euclidean error over 14 keypoints against the dense reference. \emph{Local} removes pelvis translation (without heading alignment); \emph{Global} uses world coordinates.
    \item \textbf{XY (m) / Yaw ($^\circ$):} Frame-level planar pelvis distance and shortest world-frame heading error against the dense reference.
    \item \textbf{AR (\%):} Command-level arrival rate adapted from S-BFM~\cite{sbfm}. A waypoint is reached if any step from issuance to 0.3\,s past its deadline satisfies $\text{Local}\le 20$\,cm, $\text{Global}\le 50$\,cm, $\text{XY}\le 0.5$\,m, and $\text{Yaw}\le 45^\circ$ relative to that target; commands whose window extends past clip end are excluded.
    \item \textbf{Aggregation:} Errors are pooled over frames and AR over eligible commands; \emph{Fixed} averages the five constant-$\Delta t$ runs equally, while \emph{Mixed} evaluates a single shared-seed random-interval schedule.
    \item \textbf{Command interfaces:} \modelname commands 5 links (pelvis, wrist-yaw, and ankle-roll links) or all 14 links with time-to-target, expressed relative to the simulated pelvis pose. METATRON G1~\cite{metatron} uses its native 5-target interface (torso, hands, ankles) with world-frame goals and deadline.
\end{itemize}

%% file: sections/09_policy_interface.tex
\section{Policy Inputs and Outputs}
\label{app:policy-io}

At each control step, the policy receives proprioceptive history and a timed reference command (Table~\ref{tab:policy-io}). The sparse prior encodes them into a command latent conditioning the MoE flow decoder (Table~\ref{tab:policy-encoder}) to generate a 5-step state--action trajectory; the robot executes the first action (Table~\ref{tab:policy-output}). All tensor shapes omit the batch dimension.

\begin{table}[!ht]
\centering
\caption{External policy inputs at a single control step.}
\label{tab:policy-io}
\small
\begin{tabularx}{\linewidth}{@{}p{0.21\linewidth}p{0.14\linewidth}L@{}}
\toprule
Input & Shape & Channels \\
\midrule
Robot history & $10\times93$ & Per frame (oldest first): projected gravity $\mathbf{g}$ (3), root angular velocity $\boldsymbol{\omega}_{\mathrm{root}}$ (3), joint offset $\mathbf{q}-\mathbf{q}_0$ (29), joint velocity $\dot{\mathbf{q}}$ (29), and previous action $\mathbf{a}_{t-1}$ (29). \\
Reference command & $5\times142$ & Per slot: 14 link poses $\{(\hat{\mathbf{p}}_b, \hat{\mathbf{R}}_b^{6\mathrm{D}})\}_{b\in\mathcal{B}}$ (126; 3D pos.\ + 6D rot.), visibility mask $\mathbf{m}\in\{0,1\}^{14}$ (14), time to target $\Delta\tau$ (1), and validity flag $v\in\{0,1\}$ (1); hidden links zero-masked. \\
\bottomrule
\end{tabularx}
\end{table}

\begin{table}[!ht]
\centering
\caption{Internal conditioning representations for the flow decoder.}
\label{tab:policy-encoder}
\small
\begin{tabularx}{\linewidth}{@{}p{0.21\linewidth}p{0.14\linewidth}L@{}}
\toprule
Representation & Shape & Construction \\
\midrule
Prior tokens & $16\times512$ & Width-512 linear projections of the current observation (newest frame), 10 history frames with temporal ages, and 5 command slots, concatenated in order. \\
Command latent & $64$ & Predicted by the current token as a diagonal Gaussian $(\boldsymbol{\mu}, \log\boldsymbol{\sigma}^2)$ (64 each); deterministic inference passes $\mathbf{z} = \boldsymbol{\mu}/\|\boldsymbol{\mu}\|_2$ to the decoder. \\
Decoder input & $2{,}135$ & Concatenation of normalized history (930), command latent $\mathbf{z}$ (64), noisy state--action sample $\mathbf{x}_u$ ($5\times228$), and flow time $u\in[0,1]$ (1). \\
\bottomrule
\end{tabularx}
\end{table}

\begin{table}[!ht]
\centering
\caption{Generated state--action trajectory and executed motor command.}
\label{tab:policy-output}
\small
\begin{tabularx}{\linewidth}{@{}p{0.21\linewidth}p{0.14\linewidth}L@{}}
\toprule
Output & Shape & Meaning \\
\midrule
State sequence & $5\times199$ & Per step: root pose $(\mathbf{p}_{\mathrm{root}}, \mathbf{R}_{\mathrm{root}}^{6\mathrm{D}})$ (9), root velocities $(\mathbf{v}_{\mathrm{root}}, \boldsymbol{\omega}_{\mathrm{root}})$ (6), joint state $(\mathbf{q}, \dot{\mathbf{q}})$ (58), and 14 link poses $\{(\mathbf{p}_b, \mathbf{R}_b^{6\mathrm{D}})\}_{b\in\mathcal{B}}$ (126). \\
Action sequence & $5\times29$ & Normalized joint targets $\mathbf{a}_{t:t+4}$ paired with preceding states; flow integration and denormalization yield 5 state--action pairs (width 228). \\
Executed action & $29$ & First action mapped to joint targets $\mathbf{q}_t^{\mathrm{des}} = \mathbf{q}_0 + \mathbf{s}_a \odot \mathbf{a}_t$ via default angles $\mathbf{q}_0$ and scales $\mathbf{s}_a$; replanned every 20\,ms. \\
\bottomrule
\end{tabularx}
\end{table}
\FloatBarrier

\paragraph{Coordinate frames and timing.}
History ages span 0.18\,s to 0\,s and each goal slot stores remaining time $\Delta\tau$ in seconds; invalid or all-masked slots are masked in self-attention. Reference poses are expressed relative to a chosen pelvis origin and the robot's current orientation: during training, the simulated pelvis position with localization noise in data collection, and the world-model-predicted pelvis position in imagined rollouts. At evaluation, HumanTracker uses the reference pelvis position rotated by the simulated pelvis orientation, whereas MotionChallenge follows a proprioception-only evaluation interface (without external global position or root linear velocity inputs) and uses a step-indexed deterministic flow-noise schedule. Generated root poses, link poses, and root velocities are expressed in the pelvis frame at horizon start, with the first pair at the current step.
\FloatBarrier

%% file: sections/10_reward_terms.tex
\section{Tracking Rewards and Control Penalties}
\label{app:rewards}

\subsection{Tracking and Survival Rewards}
\label{app:rewards-tracking}

Dense and sparse command inputs use the same reward terms, weights, and scales. Tracking rewards are evaluated against the complete reference motion in both cases.

Let $\mathcal{B}$ ($|\mathcal{B}|=14$) and $\mathcal{L}$ ($|\mathcal{L}|=5$) denote the tracked links and pelvis-frame landmarks, respectively. Hat quantities ($\hat{\cdot}$) denote the full-body reference, superscript $\mathrm{rel}$ denotes poses in each body's pelvis frame ($\mathbf{R}_b^{\mathrm{rel}} = \mathbf{R}_{\mathrm{pelvis}}^{\top}\mathbf{R}_b$), $\theta(\mathbf{R}, \hat{\mathbf{R}}) = \|\log(\mathbf{R}^{\top}\hat{\mathbf{R}})\|_2$ is the geodesic angle on $\mathrm{SO}(3)$, and $\mathbb{I}[\cdot]$ is the indicator function, as summarized in Table~\ref{tab:tracking-rewards}.

\begin{table}[!ht]
\centering
\caption{Positive tracking and survival reward terms per control step. Each scale $\sigma$ is in the units of the underlying tracking error prior to squaring.}
\label{tab:tracking-rewards}
\small
\renewcommand{\arraystretch}{1.38}
\begin{tabular*}{\linewidth}{@{\extracolsep{\fill}}llrr@{}}
\toprule
Term & Expression & Weight & Scale $\sigma$ \\
\midrule
World position & $\exp\!\left(-\frac{1}{|\mathcal{B}|\sigma^2}\sum_{b\in\mathcal{B}} \|\mathbf{p}_b - \hat{\mathbf{p}}_b\|_2^2\right)$ & $1.0$ & $0.30$\,m \\
World orientation & $\exp\!\left(-\frac{1}{|\mathcal{B}|\sigma^2}\sum_{b\in\mathcal{B}} \theta(\mathbf{R}_b, \hat{\mathbf{R}}_b)^2\right)$ & $1.0$ & $0.40$\,rad \\
Relative orientation & $\exp\!\left(-\frac{1}{|\mathcal{B}|\sigma^2}\sum_{b\in\mathcal{B}} \theta(\mathbf{R}_b^{\mathrm{rel}}, \hat{\mathbf{R}}_b^{\mathrm{rel}})^2\right)$ & $1.0$ & $0.40$\,rad \\
Linear velocity & $\exp\!\left(-\frac{1}{|\mathcal{B}|\sigma^2}\sum_{b\in\mathcal{B}} \|\mathbf{v}_b - \hat{\mathbf{v}}_b\|_2^2\right)$ & $1.0$ & $1.00$\,m/s \\
Angular velocity & $\exp\!\left(-\frac{1}{|\mathcal{B}|\sigma^2}\sum_{b\in\mathcal{B}} \|\boldsymbol{\omega}_b - \hat{\boldsymbol{\omega}}_b\|_2^2\right)$ & $1.0$ & $3.14$\,rad/s \\
Five-point position & $\exp\!\left(-\frac{1}{|\mathcal{L}|\sigma^2}\sum_{\ell\in\mathcal{L}} \|\mathbf{p}_\ell^{\mathrm{rel}} - \hat{\mathbf{p}}_\ell^{\mathrm{rel}}\|_2^2\right)$ & $1.0$ & $0.10$\,m \\
Pelvis height & $\exp\!\left(-\frac{(z_{\mathrm{pelvis}} - \hat{z}_{\mathrm{pelvis}})^2}{\sigma^2}\right)$ & $0.5$ & $0.30$\,m \\
Survival & $\mathbb{I}[\neg\,\mathrm{fail}]$ & $1.0$ & --- \\
\bottomrule
\end{tabular*}
\end{table}

The 14 tracked links $\mathcal{B}$ comprise the pelvis, torso, and bilateral hips (roll), knees, ankles (roll), shoulders (roll), elbows, and wrists (yaw). The five landmarks $\mathcal{L}$ consist of both wrists, both ankles, and a point $0.50$\,m along the torso's local positive $z$-axis.

\subsection{Control Penalties and Episode Termination}
\label{app:control-penalties}

Control penalties regularize action smoothness, joint limits, and body contacts (Table~\ref{tab:control-penalties}), where $\mathbf{a}_t \in \mathbb{R}^{29}$ is the normalized action at step $t$, $[x]_+ = \max(x, 0)$, $[q_j^{\min}, q_j^{\max}]$ are soft limits of joint $j$, $\mathcal{C}$ is the monitored bodies excluding ankle-roll and wrist-yaw links, and $\mathbf{f}_{c,\tau}$ is the contact force over sensor history $\mathcal{H}$.

\begin{table}[!ht]
\centering
\caption{Control penalty terms and evaluation scope. Weights multiply the raw expressions below.}
\label{tab:control-penalties}
\small
\renewcommand{\arraystretch}{1.35}
\begin{tabular*}{\linewidth}{@{\extracolsep{\fill}}llrr@{}}
\toprule
Term & Expression & Weight & Evaluated in \\
\midrule
Action change & $\|\mathbf{a}_t - \mathbf{a}_{t-1}\|_2^2$ & $-0.1$ & Simulator and world model \\
Joint limits & $\sum_{j=1}^{29}\big([q_j^{\min} - q_j]_+ + [q_j - q_j^{\max}]_+\big)$ & $-10.0$ & Simulator and world model \\
Undesired contact & $\sum_{c\in\mathcal{C}} \mathbb{I}\!\left[\max_{\tau\in\mathcal{H}} \|\mathbf{f}_{c,\tau}\|_2 > 1\,\mathrm{N}\right]$ & $-0.1$ & Simulator only \\
\bottomrule
\end{tabular*}
\end{table}

The total per-step reward is scaled by the control interval $\Delta t = 0.02$\,s:
\begin{equation}
r_t = \Delta t \left(\sum_{i} w_i r_{i,t} + \sum_{k} w_k c_{k,t}\right),
\end{equation}
where $r_{i,t}$ and $c_{k,t}$ denote the tracking/survival terms (Table~\ref{tab:tracking-rewards}) and penalty terms (Table~\ref{tab:control-penalties}), respectively. In simulation, an episode terminates early on tracking failure ($\mathrm{fail}$) if $\max_{b\in\mathcal{B}} \|\mathbf{p}_b - \hat{\mathbf{p}}_b\|_2 > 0.50$\,m, or at the 500-step (10\,s) horizon; this rule is distinct from the benchmark criteria in Appendix~\ref{app:benchmark-metrics}.
\FloatBarrier

%% file: sections/11_training_schedule.tex
\section{Training Schedule and Data Collection}
\label{app:training-schedule}

Table~\ref{tab:training-budget} details the compute budget for the default six-layer-encoder, five-step-flow policy trained on eight NVIDIA RTX 5090 GPUs. Supervised training comprises behavior cloning and four DAgger rounds, followed by world-model reinforcement learning. Elapsed time includes data collection, preprocessing, and optimization within each stage. GPU-hours equal elapsed hours multiplied by eight, with totals computed before rounding.

\begin{table}[!ht]
\centering
\caption{Compute budget for a full training run. The DAgger row aggregates all four rounds.}
\label{tab:training-budget}
\small
\begin{tabular*}{\linewidth}{@{\extracolsep{\fill}}lrrr@{}}
\toprule
Training stage & Policy updates & Elapsed hours & GPU-hours \\
\midrule
Behavior cloning & 40,000 & 8.6 & 69 \\
DAgger & 40,000 & 10.5 & 84 \\
Model-based reinforcement learning & 50,000 & 103.2 & 826 \\
\midrule
Full training run & 130,000 & 122.4 & 979 \\
\bottomrule
\end{tabular*}
\end{table}

\subsection{Behavior Cloning and Dataset Aggregation}
The dense teacher tracks motions sampled from the corpus in Section~\ref{sec:training-data}. Behavior cloning uses 20,172,254 training frames and 365,949 validation frames. This stage takes 8.6\,h, comprising 1.0\,h for data collection, 0.5\,h for preprocessing, and 7.1\,h for supervised optimization.

In each DAgger round, the sparse student executes for 10--50 control steps before the dense teacher takes over for 20 consecutive steps to provide corrective demonstrations. Each round contributes roughly one million newly labeled frames and takes approximately 2.6\,h in total (covering data collection and 10,000 supervised updates). Across all four rounds, we aggregated 4,000,280 additional expert-labeled frames, expanding the supervised dataset to 24,172,534 frames.

\subsection{World-Model Optimization}
We initialize world-model reinforcement learning from the final DAgger checkpoint and collect fresh on-policy transitions across 16,384 parallel simulator environments (2,048 per GPU). After warming up the replay buffer with 64 control steps per environment, we alternate between two and three environment steps per actor update.

The dynamics model is initially trained for 5,000 optimizer steps and updated for 10 steps per subsequent cycle using 8-step prediction horizons. Each actor update samples 1,024 replay states per GPU and backpropagates through imagined trajectories of up to 32 steps. The reported 103.2-hour duration covers replay warm-up, simulator rollouts, dynamics fitting, and all 50,000 policy updates.
\FloatBarrier